\documentclass[10pt,twocolumn,letterpaper]{article}

\usepackage{cvpr}              

\usepackage{graphicx}
\usepackage{amsmath}
\usepackage{amssymb}
\usepackage{booktabs}
\usepackage{multirow}
\usepackage{booktabs}
\usepackage{tabularx}
\usepackage{makecell}
\usepackage{multirow}
\usepackage{makecell}
\usepackage{marvosym}
\usepackage{capt-of}
\usepackage{comment}
\usepackage{multicol}
\usepackage{algorithm}
\usepackage{algcompatible}
\usepackage{algorithmicx}
\usepackage{algpseudocode}
\usepackage{subcaption}
\usepackage[accsupp]{axessibility}
\usepackage[dvipsnames]{xcolor}

\newcolumntype{Y}{>{\centering\arraybackslash}X}

\usepackage{stackengine}

\DeclareMathOperator{\E}{\mathbb{E}}
\algnewcommand{\LeftComment}[1]{\Statex \(\triangleright\) #1}

\usepackage[dvipsnames]{xcolor}

\newcommand{\keyword}[1]{{\noindent\textbf{#1}}}

\definecolor{cvprblue}{rgb}{0.21,0.49,0.74}
\usepackage[pagebackref,breaklinks,colorlinks,citecolor=cvprblue]{hyperref}

\def\paperID{321} 
\def\confName{CVPR}
\def\confYear{2024}

\title{InterHandGen: Two-Hand Interaction Generation \\via Cascaded Reverse Diffusion}
\vspace{-2\baselineskip}
\author{Jihyun Lee$^1$
\quad
Shunsuke Saito$^2$
\quad
Giljoo Nam$^2$
\quad
Minhyuk Sung$^1$
\quad
Tae-Kyun Kim$^{1,3}$\\
$^1$ KAIST \quad $^2$ Codec Avatars Lab, Meta \quad $^3$ Imperial College London\\
\small{\url{https://jyunlee.github.io/projects/interhandgen}}}

\vspace{-2\baselineskip}

\begin{document}

\twocolumn[{%
\maketitle

\renewcommand\twocolumn[1][]{#1}%
\maketitle
\begin{center}
\centering
\captionsetup{type=figure}
\vspace{-\baselineskip}
\includegraphics[width=\textwidth]{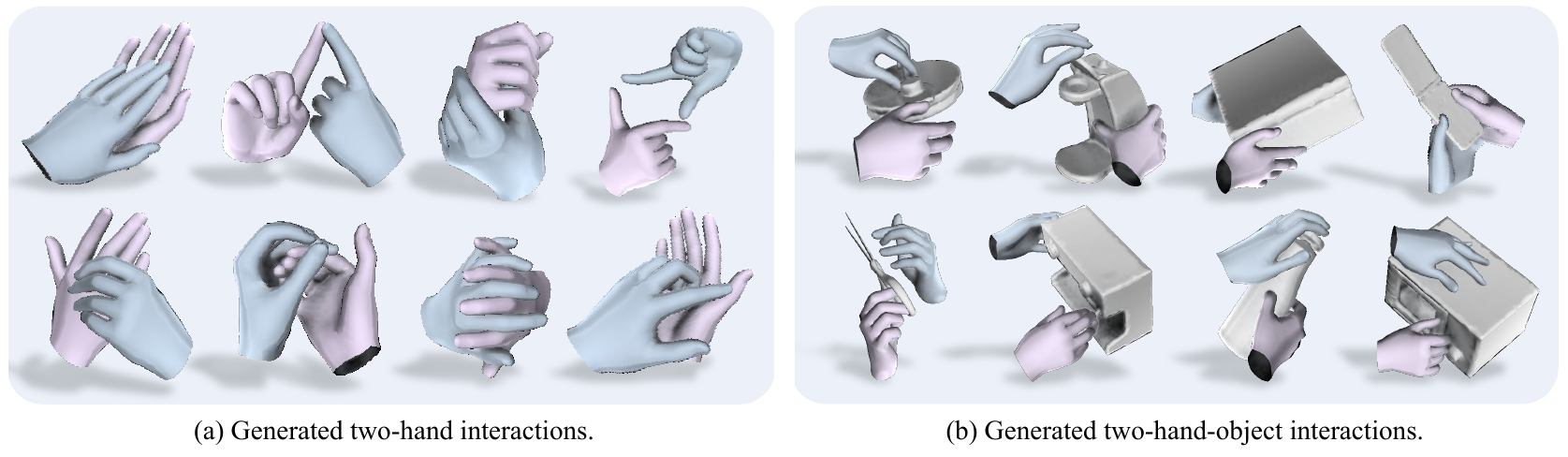} 
\vspace{-1.6\baselineskip}
\caption{\textbf{Two-hand synthesis with InterHandGen.} We propose InterHandGen, an approach to generate two-hand interactions with or without an object using a novel cascaded diffusion. To enable high-fidelity and diverse sampling, we decompose the modeling of joint distribution into the modeling of factored unconditional and conditional single-hand distributions.}
\label{fig:teaser_image}
\vspace{0.5\baselineskip}
\end{center}
}]

\begin{abstract}
\vspace{-0.5\baselineskip}

We present InterHandGen, a novel framework that learns the generative prior of two-hand interaction.
Sampling from our model yields plausible and diverse two-hand shapes in close interaction with or without an object. 
Our prior can be incorporated into any optimization or learning methods to reduce ambiguity in an ill-posed setup. 
Our key observation is that directly modeling the joint distribution of multiple instances imposes high learning complexity due to its combinatorial nature.
%
Thus, we propose to decompose the modeling of joint distribution into the modeling of factored unconditional and conditional \textbf{single} instance distribution. 
In particular, we introduce a diffusion model that learns the single-hand distribution unconditional and conditional to another hand via conditioning dropout.
For sampling, we combine anti-penetration and classifier-free guidance to enable plausible generation.
Furthermore, we establish the rigorous evaluation protocol of two-hand synthesis, where our method significantly outperforms baseline generative models in terms of plausibility and diversity. 
We also demonstrate that our diffusion prior can boost the performance of two-hand reconstruction from monocular in-the-wild images, achieving new state-of-the-art accuracy.

\end{abstract}    
\section{Introduction}
\label{sec:intro}

Two-hand interaction is widely involved in our daily lives. We coordinate our hands closely together when clasping, praying, stretching, or engaging in social interactions. Modeling and understanding two-hand interactions are thus crucial for applications that require capturing human behaviors, such as augmented or virtual reality (AR/VR) and human-computer interaction (HCI). Highlighting this importance, numerous research endeavors have been dedicated to interacting hands reconstruction. With the release of the large-scale two-hand interaction dataset~\cite{moon2020interhand2}, various methods~\cite{li2022interacting,zhang2021interacting,moon2020interhand2,lee2023im2hands,lee2023fourierhandflow,zuo2023reconstructing,jiang2023a2j,moon2023bringing,ren2023decoupled,rong2021monocular} have been proposed mainly for monocular two-hand reconstruction. 

The under-explored part in the current two-hand interaction literature is interacting two-hand \emph{generation}. Although there are generative models proposed for other human interaction domains (e.g., hand-object~\cite{karunratanakul2020grasping, corona2020ganhand, karunratanakul2021skeleton, jiang2021hand, turpin2022grasp, huang2023diffusion} or two-human~\cite{muller2023generative,liang2023intergen,shafir2023human} interaction), directly adapting them for two-hand interaction leads to sub-optimal generations. Compared to hand-object interaction that involves a rigid object, two hands lead to significantly more complex interactions due to the higher degree of freedom in two articulated hands. Additionally, while human-to-human body interaction is typically constrained on a shared ground plane, each joint of two hands has a full 6 DOF to allow more diverse interactions. 
Motivated by the advancement of unconstrained pose estimation leveraging a strong prior in other domains~\cite{pavlakos2019expressive,muller2023generative}, our goal is to build a highly expressive generative prior for two-hand interaction, which can be effortlessly incorporated into existing learning and optimization frameworks.



In this paper, we introduce InterHandGen, a framework that effectively learns the generative prior of two-hand interaction. The important challenge in two-hand interaction generation lies in its high data complexity caused by the combination of hand articulations. To reduce the complexity of learning such generation target, we propose to reformulate the two-hand distribution modeling into the modeling of single-hand model distribution unconditional and conditional to the other hand, such that:

\vspace{-0.2\baselineskip}
\begin{equation}
p_{\phi}(\mathbf{x}_{l}, \mathbf{x}_{r}) = p_{\phi}(\mathbf{x}_{l})\, p_{\phi}(\mathbf{x}_{r} | \mathbf{x}_{l}),
\label{eq:intro}
\end{equation}

\noindent where $p_{\phi}(\cdot)$ is the model distribution, and $\mathbf{x}_{l}$ and $\mathbf{x}_{r}$ are left and right hand shapes in interaction, respectively. By leveraging the symmetric nature of the left and right hands, we jointly learn both $p_{\phi}(\mathbf{x}_{l})$ and $p_{\phi}(\mathbf{x}_{r} | \mathbf{x}_{l})$ in the shared hand parameter domain based on MANO~\cite{romero2017embodied} model. In particular, we take a diffusion-based approach~\cite{ho2020denoising,song2021denoising} and train a single denoising diffusion model via conditioning dropout~\cite{ho2021classifier} to model both types of single-hand distribution. This way, the degree of freedom of each generation process is effectively reduced. 
Importantly, this formulation can be easily extended to two-hand and object interaction generation, by simply adding an object conditioning $\textbf{c}$ to each of the terms in Equation~\ref{eq:intro}.

In inference time, we sample one hand using the learned model $p_{\phi}(\mathbf{x}_{l})$ and the other hand conditioned on the previously sampled hand using $p_{\phi}(\mathbf{x}_{r} | \mathbf{x}_{l})$ in a cascaded manner. For conditional sampling, we use classifier-free guidance~\cite{ho2021classifier} to achieve a better balance between fidelity and diversity. To avoid sampling a physically implausible state due to penetration, we also introduce anti-penetration guidance that penalizes inter-penetration during the reverse diffusion process.
Furthermore, we show how to incorporate the learned two-hand interaction prior into any optimization or learning methods for reducing ambiguity in an ill-posed setup, inspired by Score Distillation Sampling~\cite{poole2022dreamfusion} and BUDDI~\cite{muller2023generative}. 

As there is no established benchmark for two-hand generation, we introduce a new evaluation protocol of two-hand interaction synthesis. In particular, we extend the standard metrics used for generative modeling (e.g., FID~\cite{heusel2017gans}, KID~\cite{binkowski2018demystifying}, Diversity~\cite{petrovich2021action,raab2023modi,tevet2022human}) to two-hand interaction by training a tailored feature backbone network. 
Our experiments show that our approach significantly outperforms the baseline methods on two-hand interaction generation with or without an object. We also show that our diffusion prior is useful for the downstream task of interacting two-hand reconstruction from in-the-wild images, where we set \emph{new state-of-the-art}.

Our main contributions are summarized as follows:

\begin{itemize}
    \item We propose an effective learning framework to build a generative prior of two-hand interaction. Our cascaded reverse diffusion approach shows significant improvement over baselines in terms of fidelity and diversity.
    \item Our formulation is general and can be extended to more instances. We show that our approach also achieves superior performance on two-hand interaction with objects.
    \item Our approach is a drop-in replacement for regularization in optimization or learning problems. By incorporating our prior, we achieve the state-of-the-art performance on interacting two-hand pose estimation from in-the-wild images.
    \item We provide a comprehensive analysis of two-hand generation with a newly established evaluation protocol. Our code and backbone network weights are publicly available for benchmarking future research. 
\end{itemize}
\section{Related Work}
\label{sec:related_work}

In this section, we discuss the related work on interacting two-hand reconstruction, and hand-object and two-human interaction generation. Note that the background on diffusion models can be found in Section~\ref{subsec:background}.


\keyword{Interacting two-hand reconstruction.}
Various methods have been proposed for interacting two-hand reconstruction from monocular RGB~\cite{li2022interacting,zhang2021interacting,moon2020interhand2,lee2023im2hands,lee2023fourierhandflow,zuo2023reconstructing,jiang2023a2j,moon2023bringing,ren2023decoupled,rong2021monocular}, multi-view RGB~\cite{ballan2012motion}, or depth~\cite{mueller2019real, taylor2017articulated, oikonomidis2012tracking}. To address self-similarity, self-occlusion, and complex articulations of interacting hands, the recent methods mainly exploit attention mechanism \cite{li2022interacting,moon2020interhand2,yu2023acr,zuo2023reconstructing,ren2023decoupled} and/or interaction-aware shape refinement~\cite{lee2023im2hands,ren2023decoupled,rong2021monocular,zhang2021interacting}. Recently, Zuo \emph{et al}.~\cite{zuo2023reconstructing} (which is concurrent work to ours) proposes to use a variational autoencoder (VAE)~\cite{kingma2013auto} as a prior for monocular two-hand reconstruction. While their approach is specialized for monocular image-based reconstruction using a specific network architecture, our approach can be used for any optimization and learning tasks. 
In addition, our experiments (Section~\ref{sec:experiments}) show that our diffusion-based prior significantly outperforms the vanilla VAE used in~\cite{zuo2023reconstructing} for a generation task in all metrics. 


\keyword{Hand-object interaction generation.} 
Most of the methods mainly focus on generating single-hand shapes conditioned on an object~\cite{karunratanakul2020grasping, corona2020ganhand, karunratanakul2021skeleton, jiang2021hand, turpin2022grasp, huang2023diffusion, garcia2020physics, baek2020weakly}. As the existing single-hand and object interaction datasets~\cite{chao2021dexycb, hampali2020honnotate, hasson2019learning, kwon2021h2o, liu2022hoi4d, armagan2020measuring, garcia2018first} are mostly limited to grasping~\cite{fan2023arctic}, the state-of-the-art generation methods actively leverage contact prior~\cite{jiang2021hand,liu2023contactgen,grady2021contactopt,turpin2022grasp} or physics simulators~\cite{turpin2022grasp, huang2023diffusion} to synthesize grasps that cannot be easily broken by applying external force~\cite{corona2020ganhand,huang2023diffusion,jiang2021hand}. However, in two-hand interaction, each hand can arbitrarily move by itself, so physical contact between hands does not necessarily occur. Thus, it is non-trivial to directly adapt the existing methods that heavily rely on physical priors. 
In addition, we consider the recent benchmark (ARCTIC~\cite{fan2023arctic}) on \emph{two-hand} and object interaction that captures various bimanual scenarios (e.g., opening a box, operating an espresso machine). Since ARCTIC is also not limited to dense contacts between object and both hands (e.g., grasping), our general approach outperforms the most recent method on single-hand and object interaction synthesis (ContactGen~\cite{liu2023contactgen}) extended for two-hand and object interaction generation on ARCTIC dataset.




 



\keyword{Two-human interaction synthesis.} 
More recently, a few methods for two-human interaction synthesis have been proposed. PriorMDM~\cite{shafir2023human} and InterGen~\cite{liang2023intergen} introduce diffusion models for text-driven two-human motion generation. BUDDI~\cite{muller2023generative} (which is concurrent work to ours) proposes an unconditional generation method of interacting two-human shapes. It introduces a transformer-based diffusion model to generate SMPL~\cite{loper2015smpl} parameters of two humans \emph{jointly}. In our work, we discover that directly modeling the joint distribution of two hands leads to sub-optimal generation performance due to the high data complexity. Instead, we simplify the learning process by decomposing the joint distribution into conditional and unconditional single-hand distributions and experimentally show that ours yields \emph{significantly} better generation results than BUDDI modified to synthesize two-hand interactions. 



\section{Method}
\label{sec:method}


\subsection{Preliminary}
\label{subsec:background}

\keyword{Diffusion Models.} Diffusion models (e.g., \cite{ho2020denoising, song2021denoising}) are a class of generative models that learn to recurrently transform noise $\mathbf{z}_{T} \sim \mathcal{N}(\mathbf{0}, \mathbf{I})$ into a sample from the target data distribution $\mathbf{z}_{0} \sim q(\mathbf{z}_{0})$. This denoising process is called the reverse process and can be expressed as:

\begin{equation}
p_{\phi}(\mathbf{z}_{0:T}) := p(\mathbf{z}_{T}) \prod^{T}_{t=1}p_{\phi}(\mathbf{z}_{t-1} | \mathbf{z}_{t}),
\end{equation}

\noindent where $p_{\phi}$ is a model distribution parameterized by $\phi$ and $\mathbf{z}_{1}, ..., \mathbf{z}_{T}$ are latent variables of the same dimensionality as $\mathbf{z}_{0}$. 
Conversely, the forward process models $q(\mathbf{z}_{1:T} | \mathbf{z}_{0})$ by gradually adding Gaussian noise to the data sample $\mathbf{z}_{0}$. In this process, the intermediate noisy sample $\mathbf{z}_{t}$ can be sampled as:

\vspace{-0.2\baselineskip}
\begin{equation}
\mathbf{z}_{t} = \sqrt{\alpha_{t}}\mathbf{z}_{0} + \sqrt{1-\alpha_{t}}\epsilon
\label{eq:diffusion}
\vspace{0.5\baselineskip}
\end{equation}

\noindent in variance-preserving diffusion formulation~\cite{ho2020denoising}. Here, $\epsilon \sim \mathcal{N}(\mathbf{0}, \mathbf{I})$ is a noise variable and $\alpha_{1:T} \in (0, 1]^{\textrm{T}}$ is a sequence that controls the amount of noise added at each diffusion time $t$. Given the noisy sample $\mathbf{z}_{t}$ and $t$, the diffusion model $f_{\phi}$ learns to approximate the reverse process for data generation. The diffusion model parameters $\phi$ are typically optimized to minimize $\E_{\mathbf{z}_{t}, \epsilon} \left \| \epsilon - f_{\phi}(\mathbf{z}_{t},\, t)  \right \|^{2}$~\cite{ho2020denoising} or $\E_{\mathbf{z}_{t}, \epsilon} \left \| \mathbf{z}_{0} - f_{\phi}(\mathbf{z}_{t},\, t)  \right \|^{2}$~\cite{tevet2022human,shafir2023human}. 
\noindent Note that exact formulations vary across the literature, and we kindly refer the reader to the survey papers~\cite{croitoru2023diffusion, yang2022diffusion} for a more comprehensive review of diffusion models.

\keyword{Classifier-Free Guidance (CFG)~\cite{ho2021classifier}.}
CFG is a method proposed to achieve a better trade-off between fidelity and diversity for conditional sampling using diffusion models. Instead of generating a sample using conditional score estimates only, it proposes to mix the conditional and unconditional score estimates to control a trade-off between sample fidelity and diversity:


\vspace{-0.7\baselineskip}
\begin{equation}
\tilde{f}_{\phi}(\mathbf{z}_{t}, t, \mathbf{c}) = (1 + w) f_{\phi}(\mathbf{z}_{t}, t, \mathbf{c}) - wf_{\phi}(\mathbf{z}_{t}, t, \emptyset),
\label{eq:cfg}
\end{equation}

\noindent where $\mathbf{c}$ is conditioning information and $w$ is a hyperparameter that controls the strength of the guidance. However, Equation~\ref{eq:cfg} requires training both conditional and unconditional diffusion models. To address this, Ho \emph{et al.}~\cite{ho2021classifier} introduces conditioning dropout during training, which enables the parameterization of both conditional and unconditional models using a single diffusion network. Conditioning dropout simply sets $\mathbf{c}$ to a null token $\emptyset$ with a chosen probability $p_{\mathit{uncond}}$ to jointly learn the conditional and unconditional scores during network training. Due to its ability to achieve a better balance between fidelity and diversity, CFG is used in many state-of-the-art conditional diffusion models~\cite{tevet2022human, shafir2023human, kulkarni2023nifty, nichol2022glide, brooks2023instructpix2pix, saharia2022photorealistic, poole2022dreamfusion}.


\subsection{Problem Definition and Key Formulation}
\label{subsec:problem_definition}

Our goal is to learn a distribution of 3D interacting two-hand shapes $p_{\phi}(\mathbf{x}_{l}, \mathbf{x}_{r})$ from the samples from a two-hand data distribution $q(\mathbf{x}_{l}, \mathbf{x}_{r})$. We assume a situation where one left hand $\mathbf{x}_{l}$ and one right hand $\mathbf{x}_{r}$ are interacting with each other, following the existing two-hand interaction benchmark~\cite{moon2020interhand2}. For representing each hand, we use MANO~\cite{romero2017embodied} model which is a differentiable statistical model that maps a pose parameter $\theta \in \mathbb{R}^{45}$ and a shape parameter $\beta \in \mathbb{R}^{10}$ to a hand mesh with 3D vertices $\mathbf{V} \in \mathbb{R}^{778 \times 3}$ and triangular faces $\mathbf{F} \in \mathbb{R}^{1554 \times 3}$. Based on MANO, we parameterize each hand shape as:

\begin{equation}
\mathbf{x}_{s} = [\theta_{\mathit{s}},\, \beta_{s},\, \omega_{s},\, \tau_{s}],
\label{eq:hand_representation}
\vspace{0.2\baselineskip}
\end{equation}

\noindent where $\mathbf{x}_{s} \in \mathbb{R}^{64}$ represents a 3D hand shape of side $s = \{l, r\}$, and $\theta_{s}$ and $\beta_{s}$ are the corresponding MANO pose and shape parameters. $\omega_{s} \in \mathbb{R}^{6}$ denotes the root rotation in 6D rotation representation~\cite{zhou2019continuity}, and $\tau_{s} \in \mathbb{R}^{3}$ denotes the root translation. 

To learn the distribution $p_{\phi}(\mathbf{x}_{l}, \mathbf{x}_{r})$ that captures plausible two-hand interaction states, one straightforward approach would be to directly model $p_{\phi}(\mathbf{x}_{l}, \mathbf{x}_{r})$ using a single generative network. However, we observe that the direct learning of joint two-hand distribution leads to suboptimal results, as the target distribution involves highly articulated hand shapes in close interaction, and its combinatorial nature imposes high generation complexity. To address this, our key idea is to decompose the joint two-hand distribution to model the unconditional and conditional single-hand distribution instead, such that:


\vspace{-0.2\baselineskip}
\begin{equation}
p_{\phi}(\mathbf{x}_{l}, \mathbf{x}_{r}) = p_{\phi}(\mathbf{x}_{l})\, p_{\phi}(\mathbf{x}_{r} | \mathbf{x}_{l}).
\label{eq:reformulation}
\vspace{0.4\baselineskip}
\end{equation}

\noindent Note that the joint distribution of two hands can now be represented by the distribution of a single hand on one side $p_{\phi}(\mathbf{x}_{l})$ and that on the other side $p_{\phi}(\mathbf{x}_{r})$ conditioned on $\mathbf{x}_{l}$. By decomposing the problem of learning $p_{\phi}(\mathbf{x}_{l}, \mathbf{x}_{r})$ into two sub-problems of learning unconditional and conditional single-hand distributions, we can effectively reduce the degree of freedom of each generation target. This formulation is general, and can be easily extended to two-hand and object interaction generation, by simply adding an object conditioning $\textbf{c}$ to each of the terms in Equation~\ref{eq:reformulation}. In what follows, we explain our novel parameterization of $p_{\phi}(\mathbf{x}_{l})$ and $p_{\phi}(\mathbf{x}_{r} | \mathbf{x}_{l})$ using diffusion models~\cite{ho2020denoising,song2021denoising}. Later in the experiments (Section~\ref{sec:experiments}), we also show that this simple decomposition leads to significant performance improvement in interacting two-hand generation with or without an object.




\subsection{Training}
\label{subsec:network_training}

For learning $p_{\phi}(\mathbf{x}_{l})$ and $p_{\phi}(\mathbf{x}_{r} | \mathbf{x}_{l})$ in Equation~\ref{eq:reformulation}, one straightforward approach is to separately train unconditional and conditional diffusion networks. However, there is conceptual redundancy embedded in $p_{\phi}(\mathbf{x}_{l})$ and $p_{\phi}(\mathbf{x}_{r} | \mathbf{x}_{l})$. Both distributions ultimately capture the plausible single-hand shapes, where the differences lie in (1) the side of the hand and (2) whether the distribution is unconditional or conditional. Motivated by multi-task learning~\cite{baxter2000model,thrun1995learning,zhang2021survey} that has shown that joint learning of related tasks improves both learning efficiency and accuracy by exploiting the commonalities across tasks, we also introduce a training mechanism that can jointly learn $p_{\phi}(\mathbf{x}_{l})$ and $p_{\phi}(\mathbf{x}_{r} | \mathbf{x}_{l})$ using a single diffusion network.



Regarding the difference in the side of hand, we pay attention to the observation that shape symmetry exists between left and right hands. The existing MANO model~\cite{romero2017embodied} indeed learns a unified hand model in the right-hand space, where the left-hand model is obtained by horizontally flipping the model shape space. Following MANO, we also bring all single-hand generation targets into the shared domain. Since our hand representation is already based on MANO, we follow the same mirroring transformation $\Gamma$ used in MANO~\cite{romero2017embodied} (please refer to the supplementary for details) to map the left-hand generation targets into the shared right-hand MANO parameter space for network training. In particular, our training objective can be written as:


\vspace{0.4\baselineskip}
\begin{itemize}
    \item Learning $p_{\phi}(\mathbf{x}_{r})$ from training samples of $\mathbf{x}_{r}$ and $\Gamma(\mathbf{x}_{l})$;
    \vspace{0.4\baselineskip}
    \item Learning $p_{\phi}(\mathbf{x}_{r} | \mathbf{x}_{l})$ from training samples of $(\mathbf{x}_{r}, \mathbf{x}_{l})$ and $(\Gamma(\mathbf{x}_{l}), \Gamma(\mathbf{x}_{r}))$.
\end{itemize}
\vspace{0.4\baselineskip}


\noindent 
This further augments the training data and improves generalization.
More importantly, once we normalize the hand side, our training objective becomes learning the unconditional and conditional distributions in the same right-hand MANO parameter space ($p_{\phi}(\mathbf{x}_{r})$ and $p_{\phi}(\mathbf{x}_{r} | \mathbf{x}_{l})$), rather than learning one unconditional distribution and one conditional distribution in the different hand spaces ($p_{\phi}(\mathbf{x}_{l})$ and $p_{\phi}(\mathbf{x}_{r} | \mathbf{x}_{l})$). Our new learning objective is now in the form that conditioning dropout~\cite{ho2021classifier} (Section~\ref{subsec:background}) can be directly applied to parameterize both unconditional and conditional models using a single diffusion network. 


Let our diffusion network be $D_{\phi}$ that takes a noisy hand parameter $\mathbf{x}_{t}$, a conditioning hand parameter $\mathbf{x}_{l}$ and diffusion time $t$. As shown in Algorithm~\ref{al:training}, we can train $D_{\phi}$ to enable both conditional hand generation (by taking the other hand parameter $\mathbf{x}_{l}$ as conditioning input) and unconditional hand generation (by taking $\emptyset$ as conditioning input) via conditioning dropout~\cite{ho2021classifier} (Step 3 in Algorithm~\ref{al:training}). Later in the experiments (Section~\ref{sec:experiments}), we show that training a unified diffusion network for $p_{\phi}(\mathbf{x}_{r})$ and $p_{\phi}(\mathbf{x}_{r} | \mathbf{x}_{l})$ leads to better generation results than training two separate networks. 




\vspace{0.2\baselineskip}
\begin{algorithm}
\caption{Training via conditioning hand dropout.}

\begin{algorithmic}[1]

\Require $p_{\mathit{uncond}}$: probability for conditioning dropout
\Require $\alpha_{1:T}$: diffusion noise scheduling

\Repeat
    \State Sample $(\mathbf{x}_{r}, \mathbf{x}_{l})$ from $q(\mathbf{x}_{r}, \mathbf{x}_{l})$ or $q(\Gamma(\mathbf{x}_{l}), \Gamma(\mathbf{x}_{r}))$
   
    \State $\mathbf{x}_{l} \leftarrow \emptyset$ with probability $p_{\mathit{uncond}}$ 
    
    \State $\epsilon \sim \mathcal{N}(\mathbf{0}, \mathbf{I})$
    
    \Statex \Comment{\emph{Compute diffused data at time $t$ (Equation~\ref{eq:diffusion})}\hspace{0.6cm}}

    \State $\mathbf{x}_{t} = \sqrt{\alpha_{t}}\mathbf{x}_{r} + \sqrt{1 - \alpha_{t}}\epsilon$     

    \State Take a gradient step on $\nabla_{\phi} \left \| \mathbf{x}_{r} - D_{\phi}(\mathbf{x}_{t},\, \mathbf{x}_{l},\, t))\right \|^{2}$
\Until{converged}
\end{algorithmic}
\label{al:training}
\end{algorithm}



\subsection{Inference: Cascaded Reverse Diffusion}
\label{subsec:network_inference}

After training our diffusion network, we can first sample an anchor left-hand $\mathbf{x}_{l}$ from the learned $p_{\phi}(\mathbf{x}_{r})$ after flipping the model space by $\Gamma$~\cite{romero2017embodied}. Then, we can sample an interacting right-hand conditioned on the anchor hand $\mathbf{x}_{l}$ from $p_{\phi}(\mathbf{x}_{r} | \mathbf{x}_{l})$ in the form of cascaded inference. Our overall inference procedure is described in Algorithm~\ref{al:inference}. Note that $\mathcal{E}(\cdot)$ denotes a function that computes the added noise $\epsilon$ from the diffusion model prediction~\cite{tevet2022human,shafir2023human}. In Algorithm~\ref{al:inference}, we incorporate two types of guidance into the reverse process: (1) classifier-free guidance (CFG)~\cite{ho2021classifier} to control a trade-off between fidelity and diversity in conditional sampling (Step 11 in Algorithm~\ref{al:inference}) and (2) anti-penetration guidance to avoid inter-hand penetration (Step 13 in Algorithm~\ref{al:inference}). As CFG is already discussed in Section~\ref{subsec:background}, we describe our anti-penetration guidance below.

\vspace{0.2\baselineskip}
\begin{algorithm}
\caption{Inference via cascaded hand denoising.}
\begin{algorithmic}[1]

\Require $w_{\mathit{cfg}}$: classifier-free guidance strength
\Require $w_{\mathit{pen}}$: anti-penetration guidance strength
\Require $\mathcal{L}_{\mathit{pen}}$: penetration loss function

\LeftComment{\emph{Sample anchor hand $\mathbf{x}_{l}$}}

\State $\mathbf{x}_{T} \sim \mathcal{N}(\mathbf{0}, \mathbf{I})$

\ForAll{$t$ from $T$ to 1}

    \State $\hat{\epsilon} \leftarrow  \mathcal{E}(D_{\phi}(\mathbf{x}_{t}, \emptyset, t))$ 

    \Statex \Comment{\emph{DDIM~\cite{song2021denoising} sampling}\hspace{3.9cm}}
    
    \State $\mathbf{x}_{t-1} \leftarrow \sqrt{\alpha_{t-1}}(\frac{\mathbf{x}_{t} - \sqrt{1-\alpha_{t}}\hat{\epsilon}}{\sqrt{\alpha_{t}}}) + \sqrt{1-{\alpha_{t-1}}}\hat{\epsilon}$ 

\EndFor

\State $\mathbf{x}_{l} \leftarrow \Gamma(\mathbf{x}_{0})$

\LeftComment{\emph{Sample interacting hand $\mathbf{x}_{r}$ given anchor hand $\mathbf{x}_{l}$}}

\State $\mathbf{x}_{T} \sim \mathcal{N}(\mathbf{0}, \mathbf{I})$

\ForAll{$t$ from $T$ to 1}

    \State $\hat{\epsilon}_{\mathit{uncond}} \leftarrow \mathcal{E}(D_{\phi}(\mathbf{x}_{t}, \emptyset, t))$

    \State $\hat{\epsilon}_{\mathit{cond}} \leftarrow \mathcal{E}(D_{\phi}(\mathbf{x}_{t}, \mathbf{x}_{l}, t))$

    \Statex \Comment{\emph{Classifier-free guidance~\cite{ho2021classifier}}\hspace{2.83cm}}

    \State $\hat{\epsilon} \leftarrow (1+w_{\mathit{cfg}}) \hat{\epsilon}_{\mathit{cond}} - w_{\mathit{cfg}}\hat{\epsilon}_{\mathit{uncond}}$

    \Statex \Comment{\emph{DDIM~\cite{song2021denoising} sampling}\hspace{3.9cm}}

    \State $\mathbf{x}_{t-1} \leftarrow \sqrt{\alpha_{t-1}}(\frac{\mathbf{x}_{t} - \sqrt{1-\alpha_{t}}\hat{\epsilon}}{\sqrt{\alpha_{t}}}) + \sqrt{1-{\alpha_{t-1}}}\hat{\epsilon}$ 

    \Statex \Comment{\emph{Anti-penetration guidance (Section~\ref{subsec:network_inference})}\hspace{1.32cm}}

    \State $\mathbf{x}_{t-1} \leftarrow \mathbf{x}_{t-1} - w_{\mathit{pen}} \nabla_{\mathbf{x}_{t-1}}\mathcal{L}_{\mathit{pen}}(\mathbf{x}_{t-1}, \mathbf{x}_{l})$

\EndFor

\State $\mathbf{x}_{r} \leftarrow \mathbf{x}_{0}$

\end{algorithmic}
\label{al:inference}
\end{algorithm} 

\vspace{0.2\baselineskip}

\keyword{Anti-penetration guidance.} Inspired by the existing work on diffusion guidance on image domain~\cite{bansal2023universal, lee2023syncdiffusion, dhariwal2021diffusion}, we introduce test-time guidance to avoid penetration between the generated two hands. In particular, we move the current interacting hand generation $\mathbf{x}_{t-1}$ towards the negative gradient direction of the penetration loss function $\mathcal{L}_{\mathit{pen}}$ at each denoising step (Step 13 in Algorithm~\ref{al:inference}). Let $\mathbf{V}_{t-1}, \mathbf{V}_{l} \in \mathbb{R}^{778 \times 3}$ denote mesh vertices recovered from the noisy right-hand parameter $\mathbf{x}_{t-1}$ and the conditional left-hand parameter $\mathbf{x}_{l}$ using MANO~\cite{romero2017embodied} layer. In particular, we recover these vertices from clean hand parameter estimated from $t-1$ via DDIM~\cite{song2021denoising} sampling $\frac{\mathbf{x}_{t-1} - \sqrt{1-\alpha_{t-1}}\hat{\epsilon}}{\sqrt{\alpha_{t-1}}}$ to enable more robust loss computation~\cite{bansal2023universal,lee2023syncdiffusion}. Then, our penetration loss $\mathcal{L}_{\mathit{pen}}$ is defined as:

\vspace{-\baselineskip}
\begin{equation}
\mathcal{L}_{\mathit{pen}}(\mathbf{x}_{t-1}, \mathbf{x}_{l}) = \sum_{i,\, j\, \in\, \mathcal{P}(\mathbf{x}_{t-1},\, \mathbf{x}_{l})} || \mathbf{V}_{t-1}^{i} - \mathbf{V}_{l}^{j} ||_{2},
\label{eq:penetration_loss}
\end{equation}
\vspace{-0.6\baselineskip}

\noindent which is the sum of squared distances between the penetrated vertex $\mathbf{V}_{t-1}^{i}$ in one hand and its nearest vertex $\mathbf{V}_{l}^{j}$ in the other hand. Here, $\mathcal{P}$ denotes a function that returns a set of penetrated vertex indices $(i, j)$ and is defined as:

\vspace{-0.4\baselineskip}
\begin{equation}
\mathcal{P}(\mathbf{x}_{t-1}, \mathbf{x}_{l}) = \{(i,\, j) \,| -\mathbf{n}_{j}^{\textrm{T}} \cdot (\mathbf{V}_{t-1}^{i} - \mathbf{V}_{l}^{j}) > 0\},
\label{eq:penetration_idx_func}
\end{equation}

\noindent where $j$ denotes the vertex index of $\mathbf{V}_{l}$ that is nearest to $\mathbf{V}_{t-1}^{i}$, and $\mathbf{n}_{j}$ is a normal vector at $\mathbf{V}_{l}^{j}$. This way, the amount of penetration can be approximated by projecting a vector joining the nearest vertices from the two hands onto the normal vector at the anchor hand, similar to the existing hand-object reconstruction literature~\cite{hampali2020honnotate}. 



\subsection{Generative Prior for Two-Hand Problems}
\label{subsec:prior}

We now explain how our two-hand interaction prior can be easily incorporated into any optimization or learning methods to further boost the accuracy of the downstream problems, such as monocular two-hand reconstruction. Inspired by Score Distillation Sampling (SDS)~\cite{poole2022dreamfusion} and BUDDI~\cite{muller2023generative}, we treat our pre-trained two-hand diffusion model $D_{\phi}$ as a frozen critic that regularizes the current two-hand interaction state ($\mathbf{x}_{l}, \mathbf{x}_{r}$) (e.g., predicted by a reconstruction network) to move to a higher-density region. Our diffusion-based regularization term can be written as:

\vspace{-0.5\baselineskip}
\begin{equation}
\mathcal{L}_{\mathit{reg}} = ||\, \mathcal{S}\,(D_{\phi}, \mathbf{x}_{l}, \mathbf{x}_{r}) - (\mathbf{x}_{l},\, \mathbf{x}_{r})\, ||_{2},
\label{eq:sds}
\vspace{0.2\baselineskip}
\end{equation}

\noindent where $S(\cdot, \cdot, \cdot)$ denotes a function that performs a single forward-reverse diffusion step~\cite{muller2023generative} that takes as input the current two-hand interaction state ($\mathbf{x}_{l}, \mathbf{x}_{r}$) and outputs the denoised interaction $(\hat{\mathbf{x}}_{l}, \hat{\mathbf{x}}_{r})$ estimated by $D_{\phi}$. Note that we detach the gradients of the diffusion model $D_{\phi}$ following \cite{poole2022dreamfusion,muller2023generative}. $\mathcal{L}_{\mathit{reg}}$ can be incorporated as an additional regularizer into any loss function during network training or shape optimization in a plug-and-play manner.





\vspace{0.3\baselineskip}
\begin{figure}[!h]
\begin{center}
\includegraphics[width=\columnwidth]{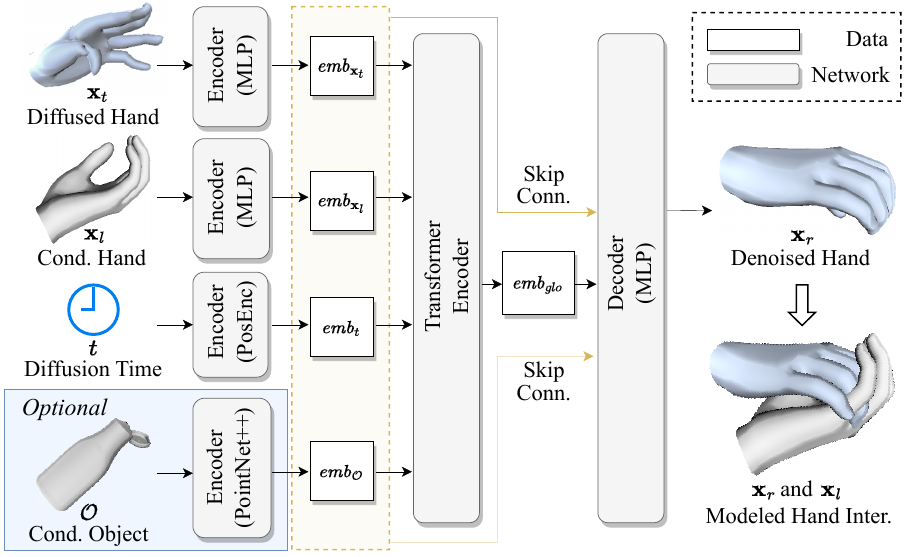}
\vspace{-1.2\baselineskip}
\caption{\textbf{Our network architecture.} We use self-attention between the embeddings of the inputs (i.e., $\mathbf{x}_{t}$ $\mathbf{x}_{l}$, $t$, and optional $\mathcal{O}$) to estimate the denoised hand parameter $\mathbf{x}_{r}$.}
\label{fig:architecture}
\vspace{-1.5\baselineskip}
\end{center}
\end{figure}

\subsection{Network Architecture}
\label{subsec:architecture}

We use a transformer-based architecture for our diffusion model $D_{\phi}$. As shown in Figure~\ref{fig:architecture}, we first use two fully connected layers with Swish activation~\cite{ramachandran2017searching} to embed the input hand and conditioning hand parameters (i.e., $\mathbf{x}_{t}$, $\mathbf{x}_{l}$). We also embed the diffusion time $t$ using Positional Encoding~\cite{ho2020denoising}. Then, we use four-headed self-attention~\cite{vaswani2017attention} to model the relationship between the input embeddings. Lastly, the updated input embeddings are flattened and fed to eight fully connected layers with ReLU~\cite{agarap2018deep} activation and skip connections to estimate the clean hand signal $\mathbf{x}_{r}$.

\keyword{Object-conditional generation.} To enable two-hand generation conditioned on an object, we can add a global object conditioning $\mathbf{c}$ to model $p_{\phi}(\mathbf{x}_{l}, \mathbf{x}_{r} | \mathbf{c}) = p_{\phi}(\mathbf{x}_{l} | \mathbf{c})\, p_{\phi}(\mathbf{x}_{r} | \mathbf{x}_{l}, \mathbf{c})$. To incorporate the object conditioning $\mathbf{c}$, we simply add a PointNet++~\cite{qi2017pointnet++}-based embedding branch (blue box in Figure~\ref{fig:architecture}) for an input object point cloud $O$. Please refer to the supplementary for more details on our architecture (e.g., layer configurations).

\section{Experiments}
\label{sec:experiments}

\begin{table*}[!t]
\centering
{ \small
\setlength{\tabcolsep}{0.2em}
\renewcommand{\arraystretch}{1.0}
\caption{\textbf{Quantitative comparisons of two-hand interaction synthesis with and without an object.} \textbf{Bold} indicates the best scores, and \underline{underline} indicates the second best scores. In both experiments, ours significantly outperforms the baselines on most of the metrics. We conduct 20 evaluations and report the average scores, where 10K samples are used in two-hand synthesis and 30K samples (3K samples per object category) are used for two-hand-object synthesis in each evaluation.}

\label{table:generation_comparisons}

\begin{subtable}[t]{\textwidth}
\caption{\textbf{Comparisons on two-hand interaction generation (Section~\ref{subsec:two-hand})}.}
\begin{tabularx}{\textwidth}{>{\centering}m{3.4cm}|Y|Y|Y|Y|Y|Y}
\toprule
Method & FHID $\downarrow$ & KHID \tiny{($\times 10^{-2}$)} \normalsize{$\downarrow$} & Diversity $\uparrow$ & Precision $\uparrow$ & Recall $\uparrow$ & PenVol \tiny{($\mathit{mm}^3$)} \normalsize{$\downarrow$} \\
\midrule
VAE~\cite{zuo2023reconstructing} & 8.18 & 6.23 & 2.32 & 0.55 & 0.02 & 7.32 \\
BUDDI*~\cite{muller2023generative} & 3.48 & 4.10 & \underline{2.71} & 0.56 & \underline{0.47} & \underline{0.82} \\
Ours w/o Decomposition  & 2.09 & 0.75 & 2.34 & \underline{0.86} & 0.35 & 3.10 \\
Ours w/o Shared Network  & \underline{1.32} & \underline{0.46} & 2.46 & \textbf{0.92} & 0.42 & 3.95 \\
\textbf{Ours} & \textbf{1.00} & \textbf{0.15} & \textbf{3.59} & \underline{0.86} & \textbf{0.85} & \textbf{0.76} \\
\bottomrule
\end{tabularx}
\label{subtable:two-hand}
\end{subtable}
\vspace{-0.4\baselineskip}

\begin{subtable}[t]{\textwidth}
\caption{\textbf{Comparisons on object-conditioned two-hand interaction generation (Section~\ref{subsec:two-hand-object})}.}
\begin{tabularx}{\textwidth}{>{\centering}m{3.4cm}|Y|Y|Y|Y|Y|Y}
\toprule
Method & FHID $\downarrow$ & KHID \tiny{($\times 10^{-1}$)} \normalsize{$\downarrow$} & Diversity $\uparrow$ & Precision $\uparrow$ & Recall $\uparrow$ & PenVol \tiny{($\mathit{mm}^3$)} \normalsize{$\downarrow$} \\
\midrule
ContactGen*~\cite{liu2023contactgen} & 22.56 & 1.58 & \underline{6.70} & 0.21 & 0.37 & 1.80\\
VAE~\cite{zuo2023reconstructing} & 21.75 & 2.12 & 5.29 & 0.60 & 0.17 & 4.98\\
BUDDI*~\cite{muller2023generative} & 22.51 & 1.35 & 6.50 & 0.28 & 0.36 & \underline{1.38}\\
Ours w/o Decomposition & 19.84 & 1.18 & 6.28 & 0.40 & \textbf{0.67} & 6.06\\
Ours w/o Shared Network & 
\underline{17.00} & \underline{0.97} & 6.15 & \textbf{0.74} & \underline{0.63} & 3.85 \\
\textbf{Ours} & \textbf{12.91} & \textbf{0.55} & \textbf{6.77} & \underline{0.71} & \textbf{0.67} & \textbf{1.33}\\
\bottomrule
\end{tabularx}
\label{subtable:two-hand-object}
\end{subtable}
}
\end{table*}

\subsection{Two-Hand Interaction Synthesis}
\label{subsec:two-hand}


\keyword{Data.}
We use InterHand2.6M~\cite{moon2020interhand2} dataset, which is the most widely used interacting two-hand dataset. Following the existing reconstruction work~\cite{li2022interacting,lee2023im2hands}, we use interacting hand (\emph{IH}) samples with \emph{valid} annotation. The resulting dataset consists of 366K training samples, 110K validation samples, and 261K test samples. 

\keyword{Baselines.}
We first consider VAE used as a two-hand prior for monocular reconstruction in \cite{zuo2023reconstructing}. We also consider BUDDI~\cite{muller2023generative}, which is a recently proposed diffusion model that \emph{jointly} generates two human parameters. We modify BUDDI to generate interacting two-hand parameters and denote the resulting model by BUDDI*. We additionally consider our method variations in which the modeling of joint distribution is not decomposed (\emph{Ours w/o Decomposition}) or separate conditional and unconditional networks are trained to model the decomposed single-hand distributions (\emph{Ours w/o Shared Network}). 
Please refer to the supplementary for the details of the baselines.

\keyword{Evaluation metrics.}
As there is no established benchmark for 3D two-hand interaction generation, we build our own evaluation protocol. Following the existing work on human pose and motion generation~\cite{raab2023modi,tevet2022human}, we extend Fréchet Inception Distance (FID)~\cite{heusel2017gans}, Kernel Inception Distance (KID)~\cite{binkowski2018demystifying}, diversity~\cite{raab2023modi,tevet2022human} and precision-recall~\cite{sajjadi2018assessing} for evaluating the generated two-hand interactions. We also report the mean inter-penetration volume in $\mathit{cm}^3$ to measure the physical plausibility. Note that FID, KID, and precision-recall are originally proposed for evaluating the feature discrepancy between the generated and the ground truth image distributions. However, there is no pre-trained feature extraction backbone for interacting two-hand shapes unlike in the image~\cite{heusel2017gans} or human motion~\cite{raab2023modi,tevet2022human} domain. To address this, we train a backbone network to extract 3D two-hand interaction features, whose network weights will be released for benchmarking future research. Inspired by FPD~\cite{shu20193d} that measures Fréchet distance of the generated 3D objects (e.g., chair, airplane) on PointNet~\cite{qi2017pointnet} feature space, we train PointNet++~\cite{qi2017pointnet++} to regress two hand poses in axis-angle representation and their relative root transformation from a 3D two-hand shape represented as a point cloud. Note that, while it is possible to extract two-hand features by specifically leveraging MANO~\cite{romero2017embodied} parameter or mesh structure, we aim to propose a more general metric for future work on two-hand interaction generation, that may not be directly reliant on the MANO model. We rename our two-hand-specific metric for FID and KID as Fréchet Hand Interaction Distance (FHID) and Kernel Hand Interaction Distance (KHID), respectively.

\begin{figure}[!h]
\begin{center}
\includegraphics[width=0.49\textwidth]{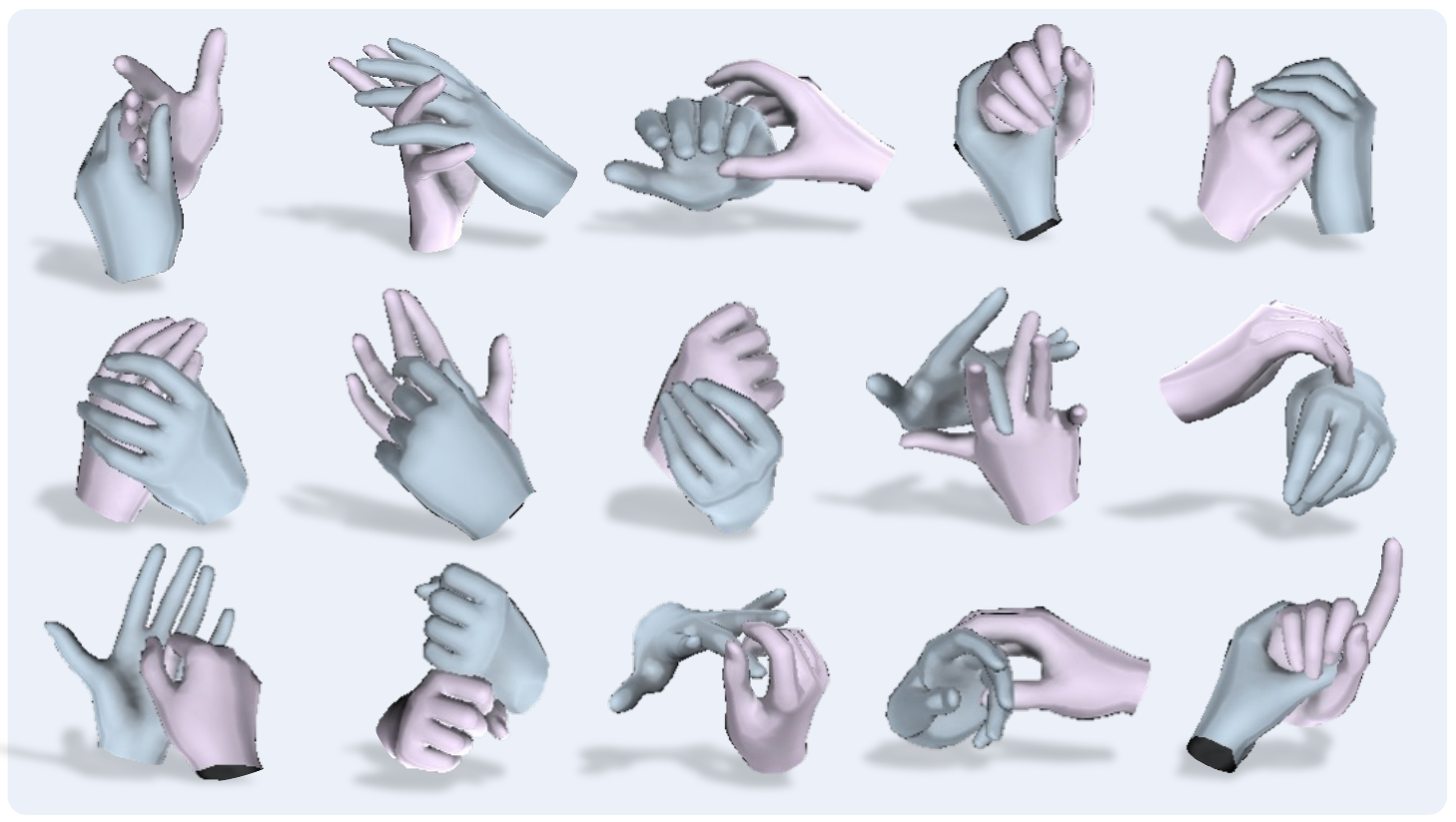} 
\caption{\textbf{Two-hand interactions synthesized by InterHandGen.} The sampled interactions are plausible and diverse.}
\label{fig:two-hand generation}
\end{center}
\vspace{-1.4\baselineskip}
\end{figure}

\keyword{Results.}
As shown in Table~\ref{subtable:two-hand}, our method significantly outperforms the baselines on most of the metrics. Especially, learning the decomposed two-hand distribution (\emph{rows 5-6}) leads to noticeable performance improvement. While \emph{Ours w/o Shared Network} (\emph{rows 5}) achieves the best precision score, our final method (\emph{rows 6}) achieves significantly better scores on the other metrics. We also notice that ours achieves high scores on both precision and recall with a good balance, while most of the baselines yield a high score on either one of them. Figure~\ref{fig:two-hand generation} qualitatively shows the sampled two-hand interactions using our method, which further demonstrates that our prior captures plausible and diverse two-hand interactions.

\vspace{0.5\baselineskip}

\subsection{Object-Conditioned Two-Hand Synthesis}
\label{subsec:two-hand-object}

\keyword{Data.}
We use the recently released ARCTIC~\cite{fan2023arctic} dataset. Unlike the existing hand-object datasets~\cite{chao2021dexycb, hampali2020honnotate, hasson2019learning, kwon2021h2o, liu2022hoi4d} that are mostly limited to single-hand grasps, ARCTIC captures diverse two-hand and object interaction scenarios, such as opening a box or operating an espresso machine. It contains 339 sequences of interaction with 10 objects. We follow the split protocol (\emph{protocol 1}) released by ARCTIC, resulting in 192K training samples, 25K validation samples, and 25K test samples. 

\keyword{Baselines.} 
We mainly consider the two-hand generation baselines from Section~\ref{subsec:two-hand} modified to additionally take an object conditioning in the same manner as our method (Section~\ref{subsec:architecture}). The baselines were further tuned to perform fair comparisons (please refer to the supplementary for details). We additionally consider ContactGen~\cite{liu2023contactgen}, which is the most recent state-of-the-art method on single-hand and object interaction synthesis. We modify ContactGen to generate two-hand interactions and denote it by ContactGen*. 


\keyword{Evaluation metrics.}
Similar to Section~\ref{subsec:two-hand}, we use FHID, KHID, diversity, precision-recall, and penetration volume. To extract two-hand interaction features relative to an object, we train a PointNet++~\cite{qi2017pointnet++} backbone network specifically for 3D two-hand and object interactions similar to Section~\ref{subsec:two-hand}. Please refer to the supplementary for the details of our backbone network. Note that we compute the metrics per object category and report the average scores. 



\begin{figure}[!h]
\begin{center}
\includegraphics[width=0.49\textwidth]{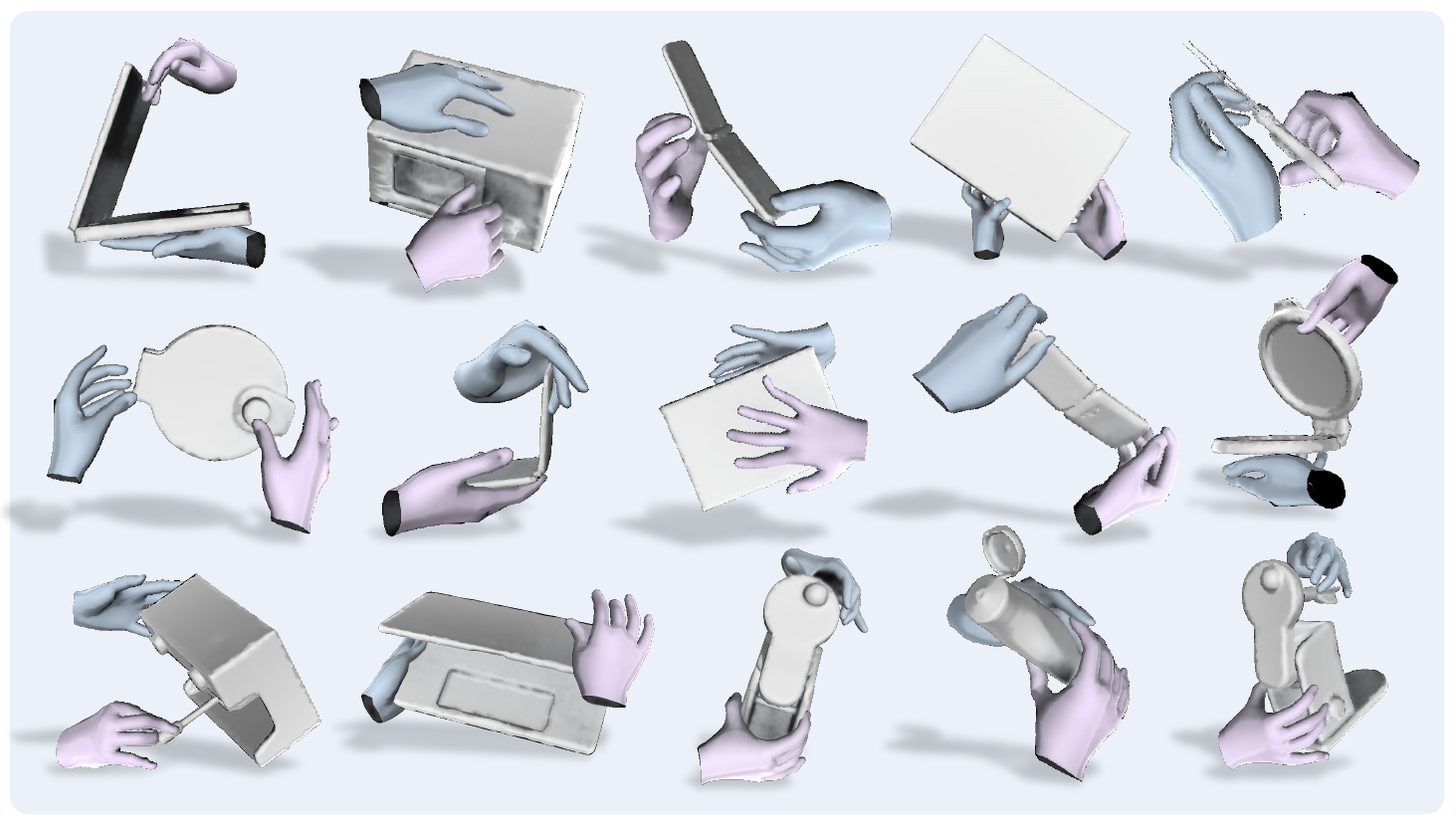} 
\vspace{-1.4\baselineskip}
\caption{\textbf{Object-conditional two-hand interaction synthesized by InterHandGen.} Ours can model plausible and diverse bimanual interactions.}
\label{fig:two-hand-obj generation}
\end{center}
\vspace{-1.5\baselineskip}
\end{figure}

\keyword{Results.}
In Table~\ref{subtable:two-hand-object}, our method is shown to outperform the baseline methods on most of the metrics by a large margin. Especially, our method yields significantly better scores on FHID and KHID. One notable observation is that ContactGen* does not achieve good performance on general two-hand and object interaction synthesis, by biasing towards heavy contact cases due to its reliance on the contact prior. In contrast, as shown in Figure~\ref {fig:two-hand-obj generation}, ours is capable of generating plausible bimanual hand interactions including loosely contacted cases.




\vspace{-0.2\baselineskip}

\subsection{Monocular Two-Hand Reconstruction}
\label{subsec:two_hand_recon}

\keyword{Baseline and Data.} 
We consider InterWild~\cite{moon2023bringing} for the baseline, which is the most recent state-of-the-art work proposed for interacting two-hand reconstruction from in-the-wild images. For network training, InterWild uses mixed-batches consisting of motion capture data with full 3D shape supervision (InterHand2.6M~\cite{moon2020interhand2}) and in-the-wild data with weak 2D keypoints supervision (MSCOCO~\cite{jin2020whole,lin2014microsoft}). In this ill-posed setup, we leverage our diffusion prior to reduce depth ambiguity.
In particular, we utilize our pre-trained two-hand diffusion model (used in Section~\ref{subsec:two-hand}) to compute the regularization term $\mathcal{L}_{\mathit{reg}}$ defined in Equation~\ref{eq:sds}. We incorporate $\mathcal{L}_{\mathit{reg}}$ into the loss function of InterWild during network training, while other baseline settings (e.g., model architecture) remain unchanged. For testing, we use InterHand2.6M~\cite{moon2020interhand2} test set and HIC~\cite{tzionas2016capturing} following the original evaluation protocol of InterWild.

\keyword{Evaluation metrics.}
We use the same metrics as in InterWild to measure the accuracy of two-hand reconstruction: Mean Per-Joint Position Error (MPJPE), Mean Per-Vertex Position Error (MPVPE), and Mean Relative-Root Position Error (MRRPE) in $\mathit{mm}$.

\keyword{Results.} As shown in Table~\ref{table:interwild}, our generative prior boosts the reconstruction accuracy of the baseline method in terms of all three metrics, \emph{setting new state-of-the-art on monocular two-hand reconstruction from in-the-wild images}. Especially, it leads to 10\% and 18\% improvements in MRRPE on InterHand2.6M and HIC datasets, respectively. These results indicate that our generative prior is effective in reducing the shape ambiguity in an ill-posed setup. We also highlight again that our pre-trained prior can be easily incorporated into the existing work in a plug-and-play manner, without a modification of the baseline architecture.

\begin{table}[!h]
\centering
{ \small
\setlength{\tabcolsep}{0.2em}
\renewcommand{\arraystretch}{1.0}
\caption{\textbf{Quantitative comparisons of interacting two-hand reconstruction from in-the-wild images.} Utilizing our generative prior can boost the two-hand reconstruction accuracy.}
\label{table:interwild}
\vspace{-0.2\baselineskip}

\begin{subtable}[t]{\columnwidth}
\caption{\textbf{Results on InterHand2.6M~\cite{moon2020interhand2}.}}
\begin{tabularx}{\columnwidth}{>{\centering}m{3.4cm}|Y|Y|Y}
\toprule
Method & MPVPE $\downarrow$ & MPJPE $\downarrow$ & MPRPE $\downarrow$\\
\midrule 
InterWild~\cite{moon2023bringing} & 13.01 & 14.83 & 29.29\\
\textbf{InterWild~\cite{moon2023bringing} + Ours} & \textbf{12.10} & \textbf{14.53} & \textbf{26.56}\\
\bottomrule
\end{tabularx}
\end{subtable}
\vspace{-0.2\baselineskip}

\begin{subtable}[t]{\columnwidth}
\caption{\textbf{Results on HIC~\cite{tzionas2016capturing}.}}
\begin{tabularx}{\columnwidth}{>{\centering}m{3.4cm}|Y|Y|Y}
\toprule
Method & MPVPE $\downarrow$ & MPJPE $\downarrow$ & MPRPE $\downarrow$\\
\midrule 
InterWild~\cite{moon2023bringing} & 15.70 & 16.17 & 31.35\\
\textbf{InterWild~\cite{moon2023bringing} + Ours} & \textbf{15.04} & \textbf{15.45} & \textbf{26.63}\\
\bottomrule
\end{tabularx}
\end{subtable}}
\end{table}




\vspace{-0.5\baselineskip}
\subsection{Ablation Study}
We perform an ablation study to investigate the effectiveness of our self-attention module (\emph{SelfAtt}), classifier-free guidance~\cite{ho2021classifier} (\emph{CFG}), and anti-penetration guidance (\emph{APG}). Table~\ref{subtable:ablation_a} compares the generated sample fidelity (measured on FHID and Precision) and diversity with respect to \emph{SelfAtt} and \emph{CFG}. It shows that using \emph{SelfAtt} improves both fidelity and diversity, while \emph{CFG} provides
a fidelity-diversity sweet spot as discussed in \cite{ho2021classifier}. In Table~\ref{subtable:ablation_b}, we compare the average penetration volume (in $\mathit{cm}^3$) and penetration distance (in $\mathit{cm}$) with and without \emph{APG}. We also measure the proximity ratio, which is the ratio of generated frames that contain close two-hand interactions (where the inter-mesh distance is below $\tau = 2cm$). It is shown that \emph{APG} significantly reduces the amount of shape penetration while not hurting the proximity ratio.


\begin{table}[!h]
\centering
{ \small
\setlength{\tabcolsep}{0.2em}
\renewcommand{\arraystretch}{1.0}
\caption{\textbf{Ablation study results.} We use the same setting as in the two-hand interaction generation experiments (Section~\ref{subsec:two-hand}).}
\label{table:ablation}
\vspace{-0.5\baselineskip}

\begin{subtable}[t]{\columnwidth}
\caption{\textbf{Comparisons on sample fidelity and diversity.} We compare to our method variations in which self-attention (\emph{Ours w/o SelfAtt}) or classifier-free guidance (\emph{Ours w/o CFG}) is not used, respectively.} 
\label{subtable:ablation_a}
\begin{tabularx}{\columnwidth}{>{\centering}m{2.4cm}|Y|Y|Y}
\toprule
Method & FHID $\downarrow$ & Precision $\uparrow$ & Diversity $\uparrow$\\
\midrule 
Ours w/o SelfAtt & 2.87 & \textbf{0.86} & 3.16\\
Ours w/o CFG & 1.12 & 0.84 & \textbf{3.61}\\
\textbf{Ours} & \textbf{1.00} & \textbf{0.86} & 3.59\\
\bottomrule
\end{tabularx}
\end{subtable}
\vspace{-0.2\baselineskip}

\begin{subtable}[t]{\columnwidth}
\caption{\textbf{Comparisons on inter-penetration.} We compare to our method variation where anti-penetration guidance is not used (\emph{Ours w/o APG}). \emph{PenVol}, \emph{PenDist}, and \emph{ProxRatio} denote penetration volume, penetration distance, and proximity ratio, respectively.}
\label{subtable:ablation_b}
\begin{tabularx}{\columnwidth}{>{\centering}m{2.4cm}|Y|Y|Y}
\toprule
Method & PenVol $\downarrow$ & PenDist $\downarrow$ & ProxRatio $\uparrow$\\
\midrule 
Ours w/o APG & 6.58 & 0.40 & \textbf{0.97}\\
\textbf{Ours} & \textbf{0.76} & \textbf{0.04} & \textbf{0.97}\\
\bottomrule
\end{tabularx}
\end{subtable}}
\end{table}




\vspace{-\baselineskip}
\section{Conclusion and Future Work}
\label{sec:conclusion}

We presented InterHandGen, a diffusion-based framework that learns the generative prior for two-hand interaction with or without an object. 
Ours provides a theoretical framework to decompose the joint distribution into a sequential modeling problem with unconditional and conditional sampling from a diffusion model. 
In particular, our experiments show that achieving both diverse and high-fidelity sampling is now possible with the proposed cascaded reverse diffusion. 
Our approach can be easily extended to more instances, and is easy to integrate into existing learning methods, setting a new state-of-the-art performance on two-hand reconstruction from in-the-wild images.


\keyword{Limitation and Future Work.}
Due to the generality of our method, the proposed diffusion prior can be jointly trained with heterogeneous datasets (i.e., a single hand only, a single hand with an object, two hands, and two hands with an object) to build a universal hand prior for all hand-related tasks. Please refer to the supplementary for more discussion. 
Other future work includes the extension to the temporal dimension and other interaction synthesis problems beyond hands (e.g., animal or human bodies). We hope that our approach will be an important stepping stone towards a unified interaction prior across categories and that it will inspire follow-up work. 


\vspace{0.5\baselineskip}
\keyword{Acknowledgements.} {\footnotesize{This work was in part supported by NST grant (CRC 21011, MSIT), KOCCA grant (R2022020028, MCST), and IITP grant (RS-2023-00228996, MSIT). M. Sung acknowledges the support of NRF grant (RS-2023-00209723) and IITP grants (2022-0-00594, RS-2023-00227592) funded by MSIT, Seoul R\&BD Program (CY230112), and grants from the DRB-KAIST SketchTheFuture Research Center, Hyundai NGV, KT, NCSOFT, and Samsung Electronics.}}

{
    \small
    \bibliographystyle{ieeenat_fullname}
    \bibliography{main}
}

\clearpage
\setcounter{page}{1}
\maketitlesupplementary

\setcounter{figure}{0}
\setcounter{table}{0}
\counterwithin{figure}{section}
\counterwithin{table}{section}

\renewcommand{\thesection}{S}
\renewcommand{\thetable}{S\arabic{table}}
\renewcommand{\thefigure}{S\arabic{figure}}

In this supplementary document, we first discuss the potential use of our method to build a universal hand prior (Section~\ref{subsec:universal_prior}) and show the additional qualitative results (Section~\ref{subsec:additional_qualitative_results}) of the experiments in the main paper. We then report additional experimental comparisons between parallel and cascaded generation approaches (Section~\ref{subsec:parallal_comparisons}). Lastly, we report the implementation details (Section~\ref{subsec:implementation_details}).


\subsection{Future Work: Universal Hand Prior}
\label{subsec:universal_prior}
Due to the generality of our method, the proposed prior can be jointly trained with heterogeneous datasets to build a universal hand prior for all hand-related problems. Recall that our method learns the decomposed hand distributions using a single diffusion network via conditioning dropout. Since our network training (Algorithm~\textcolor{red}{1} in the main paper) involves learning on both single-hand and two-hand training examples to model $p_{\phi}(\mathbf{x}_{r})$ and $p_{\phi}(\mathbf{x}_{r} | \mathbf{x}_{l})$, respectively, we can incorporate any existing single-hand datasets into the training as well. Taking a step further, we can also simultaneously apply dropout to the object condition $\mathbf{c}$ to model both object-conditional and unconditional (two-)hand distributions using a single diffusion network. Overall, our learning method based on the distribution decomposition along with conditioning dropout is naturally suited to build a multi-task prior trained with heterogeneous datasets (i.e., a single hand only, a single hand with an object, two hands, and two hands with an object).

\begin{figure}[!h]
\begin{subfigure}[!h]{\columnwidth}
\includegraphics[width=\columnwidth]{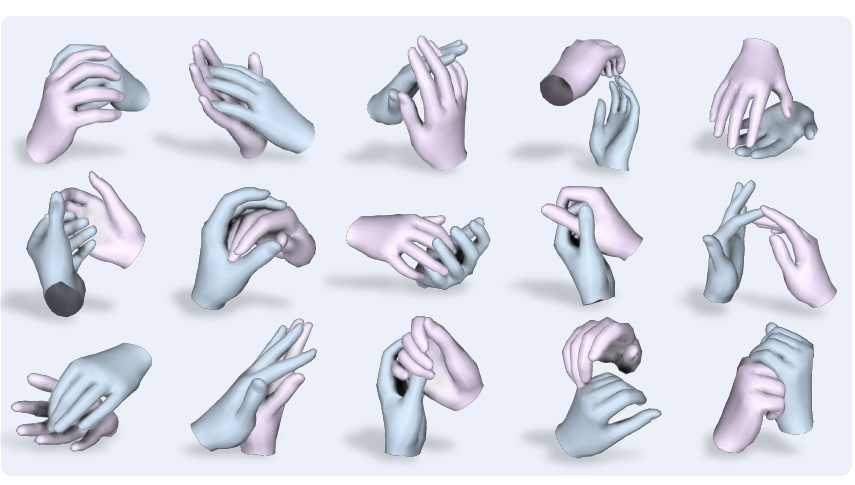} 
\vspace{-1\baselineskip}
\caption{Two-hands sampled by our prior.}
\label{subfig:univ_prior_two_hand}
\end{subfigure}

\begin{subfigure}[!h]{\columnwidth}
\includegraphics[width=\columnwidth]{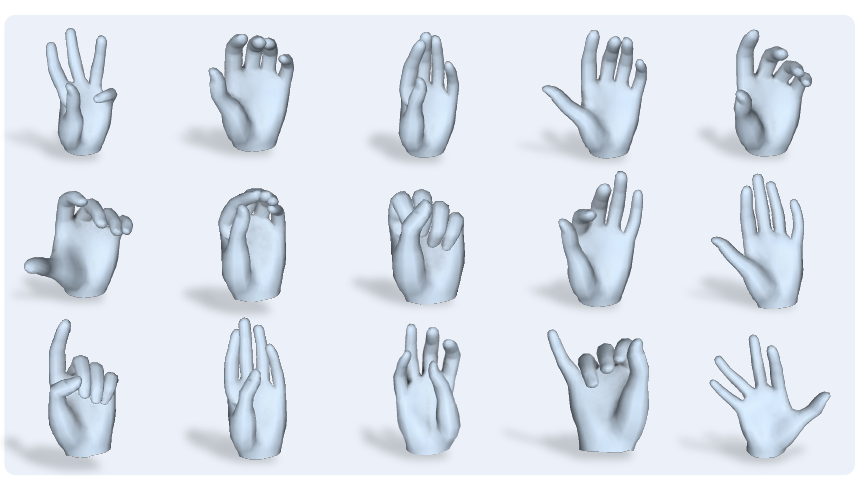} 
\vspace{-1\baselineskip}
\caption{Single-hands sampled by our prior.}
\label{subfig:univ_prior_single_hand}
\end{subfigure}

\begin{subfigure}[!h]{\columnwidth}
\includegraphics[width=\columnwidth]{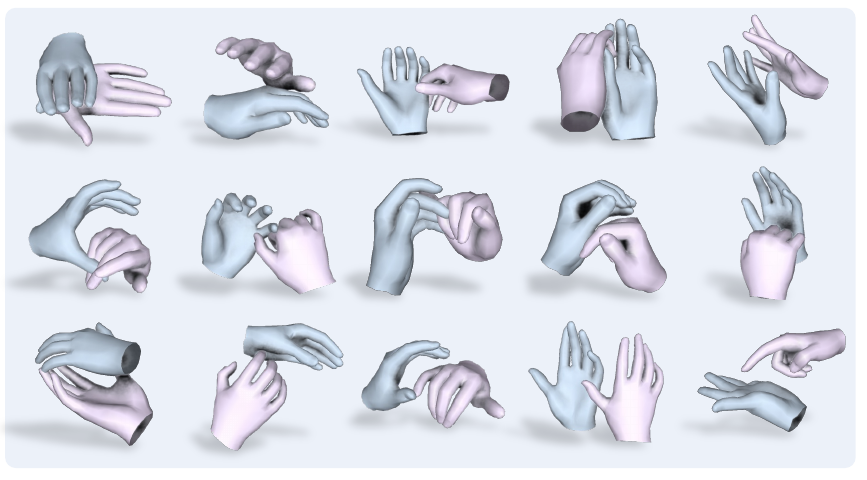} 
\vspace{-1\baselineskip}
\caption{False positive samples with respect to the manifold~\cite{sajjadi2018assessing} modeled by the prior trained on InterHand2.6M~\cite{moon2020interhand2} only.}
\label{subfig:univ_prior_false_positive}
\end{subfigure}

\caption{\textbf{Hands sampled by our prior trained on two-hand dataset~\cite{moon2020interhand2} and additional single-hand datasets~\cite{zimmermann2019freihand,zimmermann2017learning,gomez2019large,zhang2017hand}.}}
\label{fig:universal_prior_synthesis}
\vspace{-\baselineskip}
\end{figure}

While building a universal hand prior falls outside the scope of this work, we perform a toy experiment to showcase its possibility. We train our diffusion prior on two-hand dataset (InterHand2.6M~\cite{moon2020interhand2}) along with \emph{multiple single-hand datasets}~\cite{zimmermann2019freihand,zimmermann2017learning,gomez2019large,zhang2017hand} and report the qualitative examples of two-hand and single-hand synthesis in Figures~\ref{subfig:univ_prior_two_hand} and \ref{subfig:univ_prior_single_hand}, respectively. Sampling from our prior yields plausible single-hand and two-hand shapes. Importantly, this setting is shown to further boost the diversity of two-hand interaction synthesis (\textbf{from 3.59 to 4.39}) by exposing our prior to richer training examples. In Figure~\ref{subfig:univ_prior_false_positive}, we also show the generation examples that could not be sampled using the prior trained on InterHand2.6M only. In particular, we collect the generated samples that are false positive with respect to the KNN manifold~\cite{sajjadi2018assessing} modeled by the prior trained on InterHand2.6M only. As shown in the figure, these samples also model plausible two-hand interactions. One current limitation is that this universal prior does not necessarily improve the plausibility metric (e.g., FID, KID, precision) scores compared to individually trained priors. We hypothesize that existing datasets in each target domain such as InterHand2.6M~\cite{moon2020interhand2} captures only the subset of the true distributions, and individual datasets share very little with each other to bring synergy to the joint learning.
We leave building a more synergistic universal prior for future work.



\onecolumn
\subsection{Additional Qualitative Results}
\label{subsec:additional_qualitative_results}
\subsubsection{Monocular Two-Hand Reconstruction}
\begin{figure*}[!h]
\begin{center}
\vspace{-2\baselineskip}
\includegraphics[width=0.92\textwidth]{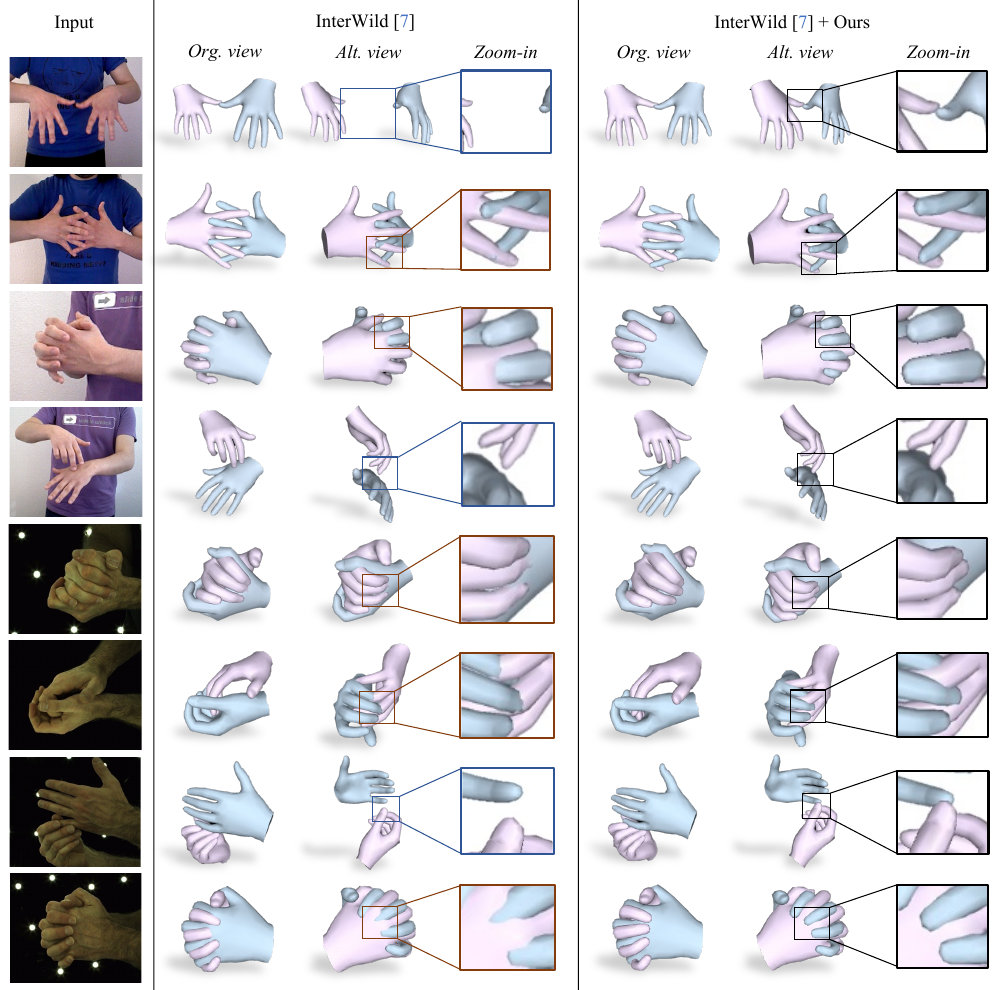} 
\caption{\textbf{Qualitative results of our monocular two-hand reconstruction experiment in Section~\textcolor{red}{4.3}.} The top four rows show results from the HIC dataset~\cite{tzionas2016capturing}, while the bottom four rows show results from the InterHand2.6M dataset~\cite{moon2020interhand2}. \textcolor{brown}{Brown boxes} highlight areas where shape penetration occurs, and \textcolor{MidnightBlue}{blue boxes} denote regions with inaccurate hand interaction (e.g., contact is absent where it should occur). Utilizing our generative prior leads to more plausible reconstructions. }
\vspace{-2\baselineskip}
\label{fig:two_hand_recon}
\end{center}
\end{figure*}

In Figure~\ref{fig:two_hand_recon}, we provide the qualitative comparison of our monocular two-hand reconstruction experiment in Section~\textcolor{red}{4.3} in the main paper. In the figure, \textcolor{brown}{brown boxes} highlight areas where shape penetration occurs, and \textcolor{MidnightBlue}{blue boxes} denote regions with inaccurate hand interaction (e.g., contact is absent where it should occur). While the baseline results of InterWild~\cite{moon2023bringing} contain several examples with penetration or inaccurate hand interaction, our approach can generate more plausible reconstructions. This indicates that leveraging our diffusion prior is effective in reducing ambiguity in an ill-posed monocular reconstruction problem. 

\subsubsection{Two-Hand Interaction Synthesis}
\begin{figure*}[!h]
\begin{center}
\includegraphics[width=0.89\textwidth]{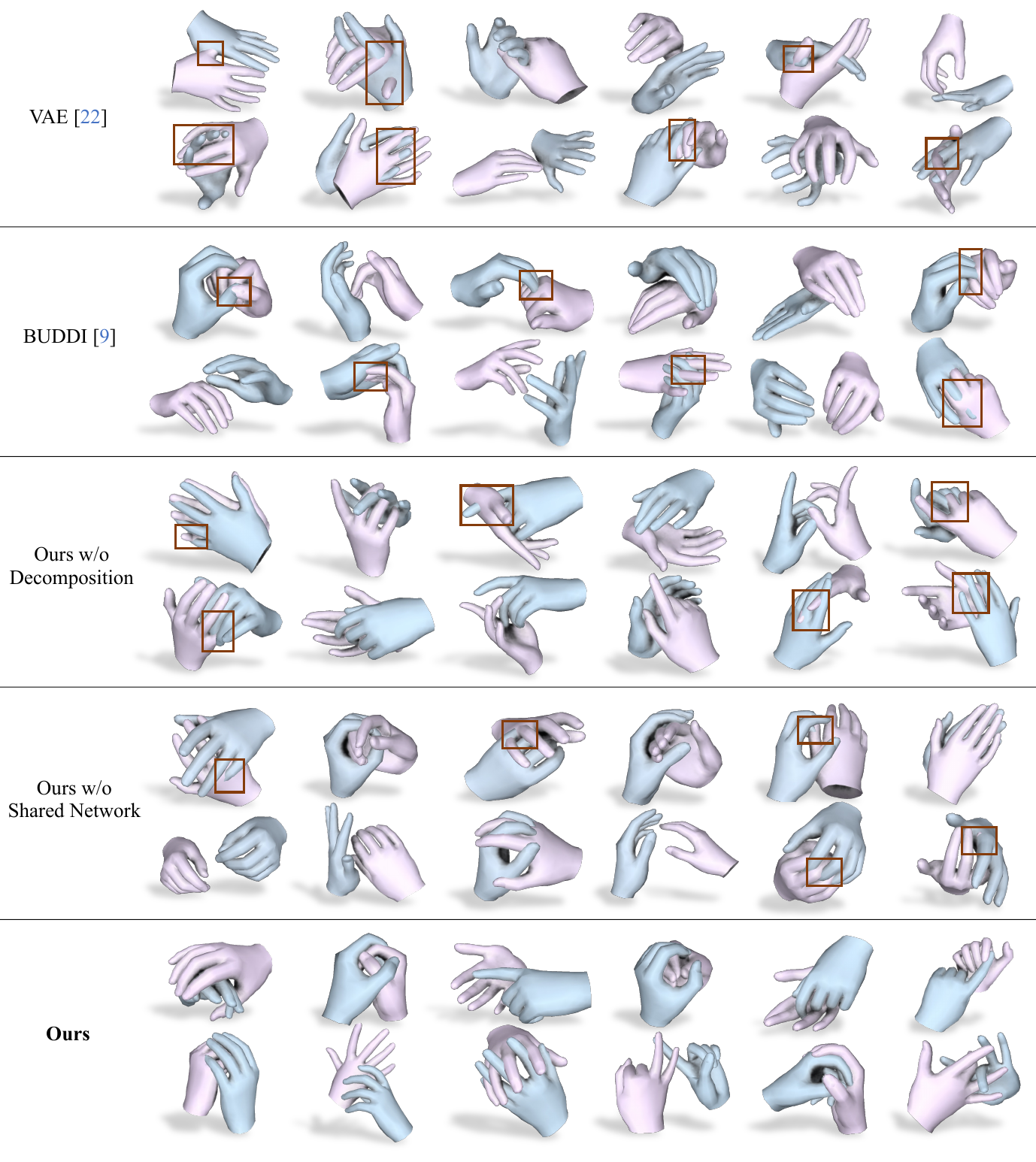} 
\vspace{-0.5\baselineskip}
\caption{\textbf{Qualitative results of two-hand interaction synthesis experiment in Section~\textcolor{red}{4.1}.} \textcolor{brown}{Brown boxes} denote regions with implausible two-hand interaction (e.g., where penetration or unnatural hand articulation occurs). Our method can produce more plausible two-hand interactions with less penetration.}
\label{fig:two_hand_gen_comp}
\vspace{-2\baselineskip}
\end{center}
\end{figure*}

In Figure~\ref{fig:two_hand_gen_comp}, we additionally show the qualitative comparison of two-hand interaction synthesis experiment in Section~\textcolor{red}{4.1} in the main paper. In the figure, \textcolor{brown}{brown boxes} denote regions with implausible two-hand interaction (e.g., where penetration or unnatural hand articulation occurs). Compared to the baselines, our method can produce more realistic two-hand interactions with less penetration. Especially, our method is shown to plausibly generate complex and tight two-hand interactions, for example, fingers of two hands crossing one another.

\subsubsection{Object-Conditioned Two-Hand Interaction Synthesis}
\begin{figure*}[!h]
\begin{center}
\includegraphics[width=0.84\textwidth]{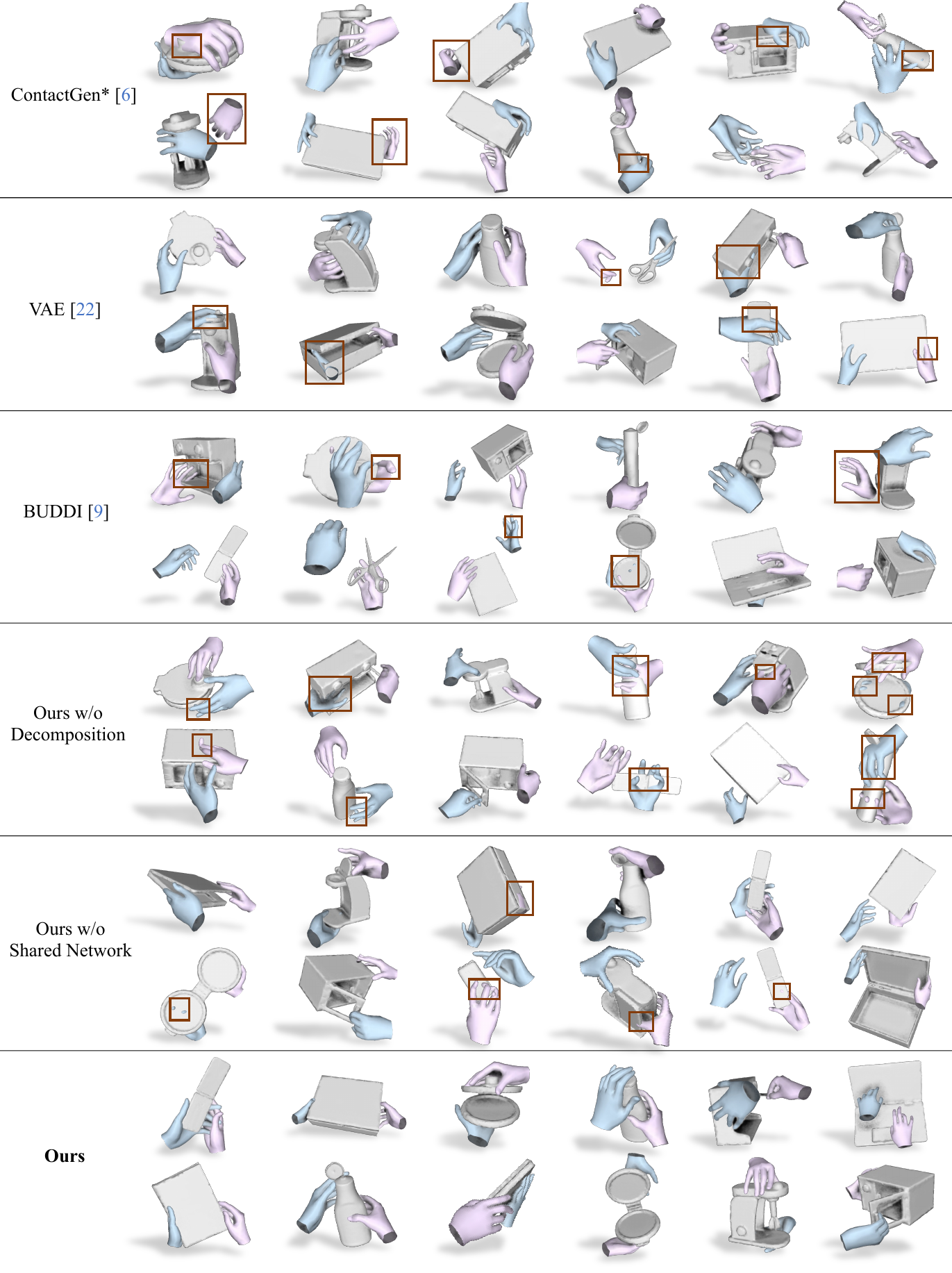} 
\vspace{-0.5\baselineskip}
\caption{\textbf{Qualitative results of two-hand interaction synthesis experiment in Section~\textcolor{red}{4.2}.} \textcolor{brown}{Brown boxes} denote implausible regions with penetration or unnatural hand articulation. Our approach can generate more realistic bimanual interactions.}
\label{fig:two_hand_obj_gen_comp}
\vspace{-3\baselineskip}
\end{center}
\end{figure*}

In Figure~\ref{fig:two_hand_obj_gen_comp}, we also report the qualitative comparisons of object-conditional two-hand synthesis experiment in Section~\textcolor{red}{4.2} in the main paper. Similar to the previous figures, \textcolor{brown}{brown boxes} denote implausible regions with penetration or unnatural hand articulation. Our approach consistently demonstrates its capability to generate more plausible two-hand interactions, that are also closely adhering to the conditioning object. 

\vspace{-0.5\baselineskip}

\twocolumn

\subsection{Parallel vs. Cascaded Generation}
\label{subsec:parallal_comparisons}

We additionally show the experimental comparisons between our cascaded generation approach and the parallel two-human generation approach of ComMDM~\cite{shafir2023human} modified for two-hand generation. Directly following \cite{shafir2023human}, we added the ComMDM communication block to two parallel single-hand diffusion networks having shared parameters. We increased the number of attention layers by one to achieve better results, while the other hyperparameters remain the same as in \cite{shafir2023human}. As shown in Tab.~\ref{table:rebuttal_intersec}, our cascaded approach leads to better generation quality due to (1) the reduced dimensionality of the generation target and (2) the conditioning on clean (rather than noisy) instances of another hand. 

\begin{table}[!h]
\centering
{ \footnotesize
\setlength{\tabcolsep}{0.2em}
\renewcommand{\arraystretch}{1.0}
\caption{\textbf{Comparisons between the parallel and cascaded generation approaches.}}
\label{table:rebuttal_intersec}
\vspace{-0.5\baselineskip}
\begin{tabularx}{\columnwidth}{>{\centering}m{3cm}|Y|Y|Y}
\toprule
Method & FHID ($\downarrow$) & Precision ($\uparrow$) & Diversity ($\uparrow$)\\
\midrule
Parallel (ComMDM~\cite{shafir2023human}) & 2.19 & 0.75 & 2.68\\
\textbf{Cascaded (Ours)} & \textbf{1.00} & \textbf{0.86} & \textbf{3.59}\\ 
\bottomrule
\end{tabularx}
}
\vspace{-0.5\baselineskip}
\end{table}

\subsection{Implementation Details}
\label{subsec:implementation_details}

We now report the implementation details for the reproducibility of the proposed method. Note that we also plan to publish our code after the review period.

\subsubsection{Evaluation Protocol}

\keyword{Two-hand feature backbone.}
We modify PointNet++~\cite{qi2017pointnet++} to regress (1) two hand poses in axis-angle representation, (2) relative root rotation in 6D rotation representation~\cite{zhou2019continuity}, and (3) relative root translation given a two-hand shape represented as a point cloud. Our network architecture mainly follows the architecture of the original PointNet++ encoder, except for the output dimension of the last fully connected layer modified to 108 (in order to match the concatenated dimension of our estimation targets). We train our network on InterHand2.6M~\cite{moon2020interhand2} dataset for 200 epochs with a batch size of 32. Other training details (e.g., learning rate, batch size) remain unchanged from the original PointNet++ model. The test MPJPE of the resulting model is 1.49$\mathit{mm}$.


\keyword{Object-conditional two-hand feature backbone.}
The network architecture and training details are the same as those of our two-hand feature backbone, except that the network regresses (1) two-hand root rotations and translations in the object-centric coordinate space (not the relative root transformation between two hands) and that (2) the object feature is additionally incorporated to estimate two-hand poses. In particular, we use the PointNet++~\cite{qi2017pointnet++} embedding module in our object-conditional diffusion model (refer to Section~\textcolor{red}{3.6}) to extract the object feature and feed it to the first fully connected layer of our two-hand pose regression network. 

\keyword{Evaluation metrics.}
We mainly follow the implementation details of the existing human pose and motion generation work~\cite{raab2023modi,tevet2022human} for computing Fréchet Distance~\cite{heusel2017gans}, Kernel Distance~\cite{binkowski2018demystifying}, diversity~\cite{raab2023modi,tevet2022human} and precision-recall~\cite{sajjadi2018assessing}. One important difference is that we adapt our own two-hand backbone network for feature extraction. For measuring penetration volume, we first voxelize two hand meshes with 1$\mathit{mm}$ grids and count the number of voxels that are occupied by both hands similar to HALO~\cite{karunratanakul2021skeleton}.


\subsubsection{Network Training and Inference}

\keyword{Training.}
We train our diffusion network for 80 epochs using an Adam optimizer with an initial learning rate of $2 \times 10^{-4}$. We additionally use a learning rate scheduler to decay the learning rate by 10\% every 20 epochs. We set the batch size as 256 and 64 for unconditional and object-conditional diffusion networks, respectively. For diffusion noise scheduling, we use linear scheduling from $\beta_{1} = 10^{-4}$ to $\beta_{T} = 0.01$~\cite{ho2020denoising}. We set the maximum value of diffusion time as $T=256$ and the probability of conditioning dropout as $p_{\mathit{uncond}} = 0.5$. Note that, for unconditional two-hand synthesis, only the relative root transformation between two hands is meaningful in modeling plausible interactions. Thus, we supervise the root transformation of the interacting hand generation ($p_{\phi}(\mathbf{x}_{r} | \mathbf{x}_{l})$) with the ground truth transformation of $\mathbf{x}_{r}$ relative to $\mathbf{x}_{l}$, while not imposing supervision to the root transformation of the anchor hand generation ($p_{\phi}(\mathbf{x}_{r})$). For object-conditional two-hand synthesis, we supervise both generation cases with the ground truth root transformations relative to the conditioning object. 

\keyword{Inference.}
For network inference, we use DDIM~\cite{song2021denoising} sampling with 32 denoising steps. We set the classifier-free guidance weight as $w_{\mathit{cfg}} = 0.1$. For anti-penetration guidance weight $w_{\mathit{pen}}$, we use a multiplicative scheduling starting from 4 at $t=0$ with a rate of 0.9. This strategy is adopted to avoid using a high weight for anti-penetration guidance in the early stages of the denoising process, where samples may still exhibit high levels of noise.

\keyword{Mirroring transformation $\Gamma$~\cite{romero2017embodied}.} We adopt the same mirroring transformation function $\Gamma(\cdot)$ used in MANO~\cite{romero2017embodied}. $\Gamma(\cdot)$ multiplies the input instance by the transformation matrix $\mathbf{T}$, which is defined as:

\vspace{-0.4\baselineskip}
\begin{equation}
\mathbf{T} = 
\begin{bmatrix}
-1 & 0 & 0\\
0 & 1 & 0\\
0 & 0 & 1
\end{bmatrix}.
\label{eq:mano_mirroring}
\end{equation}

\noindent Note that, for MANO hand shapes represented as MANO parameters, applying $\Gamma(\cdot)$ to the root rotation parameter is sufficient, as the local hand deformations are also mirrored along the MANO kinematic chain starting from the root pose (please refer to \cite{romero2017embodied} for more details on the MANO model). 

\subsubsection{Network Architecture}
\label{subsubsec:network_architecture}

\keyword{Hand embedding.}
For embedding noisy right-hand parameter $\mathbf{x}_{t} \in \mathbb{R}^{64}$ and conditioning left-hand parameter $\mathbf{x}_{l} \in \mathbb{R}^{64}$, we use two separate MLPs with the same network architecture. Each MLP consists of two fully connected layers, whose output feature dimensions are 2056 and 512, respectively. The first layer is followed by Swish activation. We denote the resulting embeddings for $\mathbf{x}_{t}$ and $\mathbf{x}_{l}$ by $\mathit{emb}_{\mathbf{x}_{t}}$, $\mathit{emb}_{\mathbf{x}_{l}}$ $\in \mathbb{R}^{512}$, respectively.

\keyword{Diffusion time embedding.}
For embedding diffusion time $t \in \mathbb{N}$, we use Sinusoidal embedding in DDPM~\cite{ho2020denoising} to extract a 512-dimensional feature. We then use an MLP (whose architecture is the same as the MLP used for hand embedding) to further extract the feature of $t$. We denote the resulting embedding for $t$ by $\mathit{emb}_{t} \in \mathbb{R}^{512}$.

\keyword{Object embedding.}
For embedding the object point cloud $\mathcal{O}$, we use a PointNet++~\cite{qi2017pointnet++}-based architecture. We modify the original PointNet++ encoder by dropping the last layer and changing the final feature dimension from 256 to 512. Other implementation details remain unchanged from \cite{qi2017pointnet++}. We denote the resulting embedding for $\mathcal{O}$ by $\mathit{emb}_{\mathcal{O}} \in \mathbb{R}^{512}$.

\keyword{Transformer encoder.}
We perform channel-wise concatenation of $\mathit{emb}_{\mathbf{x}_{t}}$, $\mathit{emb}_{\mathbf{x}_{l}}$, $\mathit{emb}_{t}$, and (optionally) $\mathit{emb}_{\mathcal{O}}$ to consider each embedding as an input token to a transformer encoder. For the architecture of the transformer encoder, we use two self-attention blocks~\cite{vaswani2017attention} with four attention heads. Each head consists of two fully connected layers, whose output feature dimensions are 2048 and 512, respectively. Each layer is followed by Layer Normalization, ReLU activation, and dropout with a rate of 0.1. After the self-attention modules, we use one fully connected layer to map the flattened output tokens into a global feature $\mathit{emb}_{\mathit{glo}} \in \mathbb{R}^{2056}$.

\keyword{Output decoder.}
We use an MLP-based decoder to estimate the clean hand parameter $\mathbf{x}_{r} \in \mathbb{R}^{64}$ from $\mathit{emb}_{\mathit{glo}}$. The MLP consists of seven fully connected layers. The output feature dimension of all layers is 2056, except for the last layer whose output dimension is 64 to model the hand parameter. Each layer (except for the last layer) is followed by ReLU activation. Note that we use skip connections for all layers, in which the input feature is concatenated with the condition embeddings (i.e., $\mathit{emb}_{\mathbf{x}_{l}}$ $\mathit{emb}_{t}$ and optional $\mathit{emb}_{\mathcal{O}}$). In the odd-numbered layers, we additionally concatenate the noisy hand embedding $\mathit{emb}_{\mathbf{x}_{t}}$ to the input feature. 

\subsubsection{Baseline Comparisons}

\keyword{Two-hand synthesis.}
For VAE~\cite{zuo2023reconstructing} and BUDDI~\cite{muller2023generative}, we use the original network architectures with minor modifications to obtain better generation results on InterHand2.6M~\cite{moon2020interhand2} dataset to perform fairer comparisons. For VAE, we empirically observed that increasing the feature dimension (from 128 to 256) and the number of encoder layers (from 4 to 5) improves the performance. For BUDDI, we increased the feature dimension of the self-attention blocks from 152 to 184 to obtain better generation results. For our method variations, we use the same implementation details except for the changes specified in Section~\textcolor{red}{4.1}.

\keyword{Object-conditional two-hand synthesis.}
For BUDDI~\cite{muller2023generative} and our method variations, we incorporate the object feature encoded by PointNet++~\cite{qi2017pointnet++} as an additional token to the transformer encoder in a similar manner to our method. For VAE~\cite{zuo2023reconstructing}, we feed the object feature as an additional input to the second layer of both the encoder and decoder, similar to HALO~\cite{karunratanakul2021skeleton}. For ContactGen~\cite{liu2023contactgen}, we extend the single-hand contact map to a two-hand contact map and optimize both hands accordingly.

%
%
%

\end{document}